\def\year{2018}\relax
\documentclass[letterpaper]{article} 
\usepackage{aaai18}  
\usepackage{times}  
\usepackage{helvet}  
\usepackage{courier}  
\usepackage{url}  
\usepackage{graphicx}  
\usepackage{amsmath}
\usepackage{amsfonts}
\usepackage{subcaption}

\begin{document}
%
\title{Learning the Enigma with Recurrent Neural Networks}
\author{Sam Greydanus\\
\texttt{sam.17@dartmouth.edu}\\
Department of Physics and Astronomy, Dartmouth College\\
Hanover, New Hampshire 03755\\
}
\maketitle

\begin{abstract}
\textbf{Recurrent neural networks (RNNs) represent the state of the art in translation, image captioning, and speech recognition. They are also capable of learning algorithmic tasks such as long addition, copying, and sorting from a set of training examples. We demonstrate that RNNs can learn decryption algorithms -- the mappings from plaintext to ciphertext -- for three polyalphabetic ciphers (Vigenere, Autokey, and Enigma). Most notably, we demonstrate that an RNN with a 3000-unit Long Short-Term Memory (LSTM) cell can learn the decryption function of the Enigma machine. We argue that our model learns efficient internal representations of these ciphers 1) by exploring activations of individual memory neurons and 2) by comparing memory usage across the three ciphers. To be clear, our work is not aimed at 'cracking' the Enigma cipher. However, we do show that our model can perform elementary cryptanalysis by running known-plaintext attacks on the Vigenere and Autokey ciphers. Our results indicate that RNNs can learn algorithmic representations of black box polyalphabetic ciphers and that these representations are useful for cryptanalysis.}
\end{abstract}

\maketitle

\section{Introduction}

Given an encrypted sequence, a key and a decryption of that sequence, can we reconstruct the decryption function? We might begin by looking for small patterns shared by the two sequences. When we find these patterns, we can piece them together to create a rough model of the unknown function. Next, we might use this model to predict the translations of other sequences. Finally, we can refine our model based on whether or not these guesses are correct.

During WWII, Allied cryptographers used this process to make sense of the Nazi Enigma cipher, eventually reconstructing the machine almost entirely from its inputs and outputs \cite{Rejewski1981HowCipher}. This achievement was the product of continuous effort by dozens of engineers and mathematicians. Cryptography has improved in the past century, but piecing together the decryption function of a black box cipher such as the Enigma is still a problem that requires expert domain knowledge and days of labor.

The process of reconstructing a cipher's inner workings is the first step of cryptanalysis. Several past works have sought to automate -- and thereby accelerate -- this process, but they generally suffer from a lack of generality (see Related Work). To address this issue, several works have discussed the connection between machine learning and cryptography \cite{Kearns1994CryptographicAutomata} \cite{Prajapat2015VariousCryptanalysis}. Early work at the confluence of these two fields has been either theoretical or limited to toy examples. We improve upon this work by introducing a general-purpose model for learning and characterizing polyalphabetic ciphers.

\begin{figure}
\includegraphics[width=.9\columnwidth]{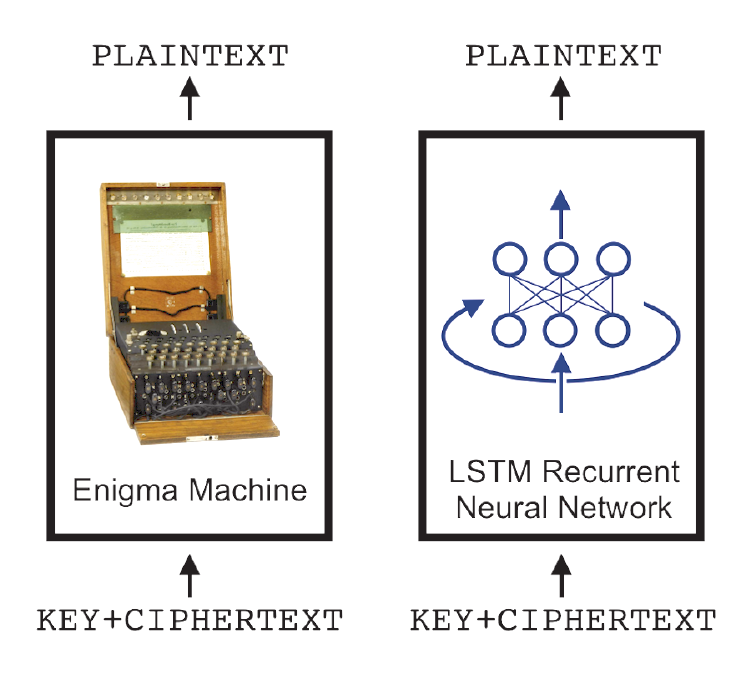}
\centering
\caption{Our LSTM-based model can learn the decryption function of the Enigma from a series of ciphertext and plaintext examples.}
\label{fig:concept}
\end{figure}
%
Our approach is to frame the decryption process as a sequence-to-sequence translation task and use a Recurrent Neural Network (RNN) -based model to learn the translation function. Unlike previous works, our model can be applied to any polyalphabetic cipher. We demonstrate its effectiveness by using the same model and hyperparameters (except for memory size) to learn three different ciphers: the Vigenere, Autokey, and Enigma. Once trained, our model performs well on 1) unseen keys and 2) ciphertext sequences much longer than those of the training set. All code is available online\footnote{\texttt{https://github.com/greydanus/crypto-rnn}}.

By visualizing the activations of our model's memory vector, we argue that it can learn efficient internal representations of ciphers. To confirm this theory, we show that the amount of memory our model needs to master each cipher scales with the cipher's degree of time-dependence. Finally, we train our model to perform known-plaintext attacks on the Vigenere and Autokey ciphers, demonstrating that these internal representations are useful tools for cryptanalysis.

To be clear, our objective was not to crack the Enigma. The Enigma was cracked nearly a century ago using fewer computations than that of a single forward pass through our model. The best techniques for cracking ciphers are hand-crafted approaches that capitalize on weaknesses of specific ciphers. This project is meant to showcase the impressive ability of RNNs to uncover information about unknown ciphers in a fully automated way.

In summary, our key contribution is that RNNs can learn algorithmic representations of complex polyalphabetic ciphers and that these representations are useful for cryptanalysis.

\section{Related work} 

Recurrent models, in particular those that use Long Short-Term Memory (LSTM), are a powerful tool for manipulating sequential data. Notable applications include state-of-the art results in handwriting recognition and generation \cite{Graves2014GeneratingNetworks}, speech recognition \cite{Graves2013SpeechNetworks}, machine translation \cite{Sutskever2014SequenceNetworks}, deep reinforcement learning \cite{Mnih2016AsynchronousLearning}, and image captioning \cite{Karpathy2015DeepDescriptions}. Like these works, we frame our application of RNNs as a sequence-to-sequence translation task. Unlike these works, our translation task requires 1) a keyphrase and 2) reconstructing a deterministic algorithm. Fortunately, a wealth of previous work has focused on using RNNs to learn algorithms.

Past work has shown that RNNs can find general solutions to algorithmic tasks. Zaremba and Sutskever \cite{Zaremba2015LearningExecute} trained LSTMs to add 9-digit numbers with 99\% accuracy using a variant of curriculum learning. Graves et. al. \cite{Graves2014NeuralMachines} compared the performance of LSTMs to Neural Turing Machines (NTMs) on a range of algorithmic tasks including sort and repeat-copy. More recently, Graves et al. \cite{Graves2016HybridMemory} introduced the Differentiable Neural Computer (DNC) and used it to solve more challenging tasks such as relational reasoning over graphs. As with these works, our work shows that RNNs can master simple algorithmic tasks. However, unlike tasks such as long addition and repeat-copy, learning the Enigma is a task that humans once found difficult.

In the 1930s, Allied nations had not yet captured a physical Enigma machine. Cryptographers such as the Polish Marian Rejewski were forced to compare plaintext, keyphrase, and ciphertext messages with each other to infer the mechanics of the machine. After several years of carefully analyzing these messages, Rejewski and the Polish Cipher Bureau were able to construct 'Enigma doubles' without ever having seen an actual Enigma machine \cite{Rejewski1981HowCipher}. This is the same problem we trained our model to solve. Like Rejewski, our model uncovers the logic of the Enigma by looking for statistical patterns in a large number of plaintext, keyphrase, and ciphertext examples. We should note, however, that Rejewski needed far less data to make the same generalizations. Later in World War II, British cryptographers led by Alan Turing helped to actually crack the Enigma. As Turing's approach capitalized on operator error and expert knowledge about the Enigma, we consider it beyond the scope of this work \cite{Sebag-Montefiore2000EnigmaCode}.

Characterizing unknown ciphers is a central problem in cryptography. The comprehensive \textit{Applied Cryptography} \cite{Schneier1996AppliedC} lists a wealth of methods, from frequency analysis to chosen-plaintext attacks. Papers such as \cite{Dawson1991BlackCiphers} offer additional methods.  While these methods can be effective under the right conditions, they do not generalize well outside certain classes of ciphers. This has led researchers to propose several machine learning-based approaches. One work \cite{Spillman2017UseCiphers} used genetic algorithms to recover the secret key of a simple substitution cipher. A review paper by Prajapat et al. \cite{Prajapat2015VariousCryptanalysis} proposes cipher classification with machine learning.

Alallayah et al. \cite{Alallayah2010AttackNeuro-Identifier} succeeded in using a small feedforward neural network to decode the Vigenere cipher. Unfortunately, they phrased the task in a cipher-specific manner that dramatically simplified the learning objective. For example, at each step, they gave the model the single keyphrase character that was necessary to perform decryption. One of the things that makes learning the Vigenere cipher difficult is choosing that keyphrase character for a particular time step. We avoid this pitfall by using an approach that generalizes well across several ciphers.

There are other interesting connections between machine learning and cryptography. Abadi and Andersen trained two convolutional neural networks (CNNs) to communicate with each other while hiding information from from a third. The authors argue that the two CNNs learned to use a simple form of encryption to selectively protect information from the eavesdropper. Another work by Ramamurthy et al. \cite{Ramamurthy2017UsingCryptography} embedded images directly into the trainable parameters of a neural network. These messages could be recovered using a second neural network. Like our work, these two works use neural networks to encrypt information. Unlike our work, the models were neither recurrent nor were they trained on existing ciphers.
\\

\section{Problem setup}

\begin{figure*}
\centering
\begin{subfigure}{\columnwidth}
  \centering
  \includegraphics[width=\columnwidth]{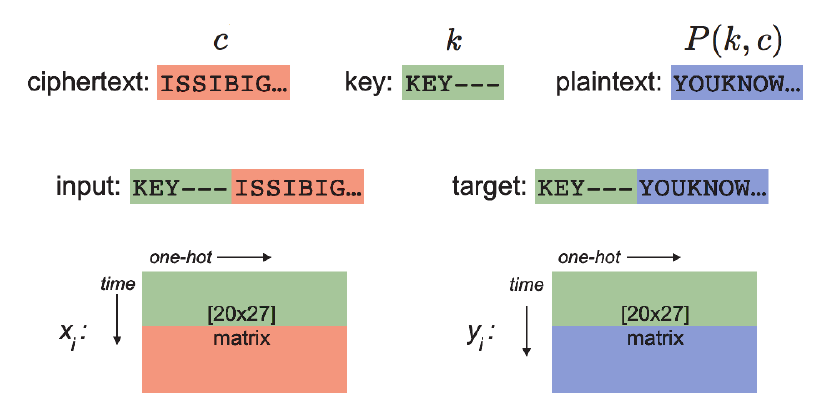}
  \caption{Data}
  \label{fig:data}
\end{subfigure}
\begin{subfigure}{\columnwidth}
  \centering
  \includegraphics[width=\columnwidth]{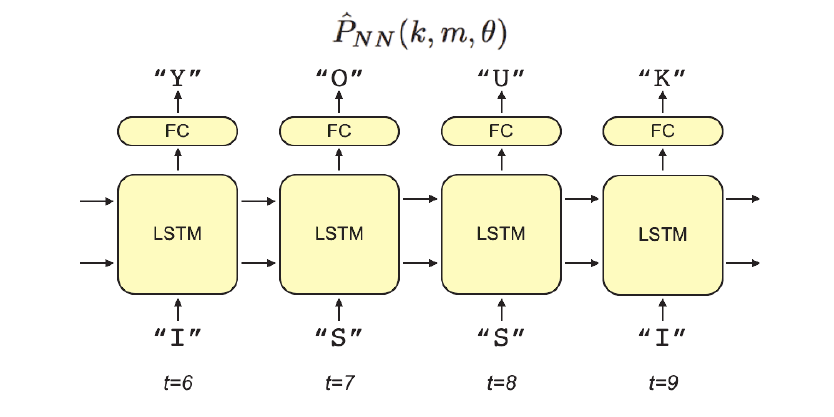}
  \caption{Model}
  \label{fig:model}
\end{subfigure}
\caption{(a) Expressing the decryption process as a sequence-to-sequence translation task. (b) Our Recurrent Neural Network (RNN) -based model unrolled for time steps 6 to 9 (FC: fully-connected layer).}
\label{fig:schema}
\end{figure*}

We consider the decryption function $m = P(k, c)$ of a generic polyalphabetic cipher where $c$ is the ciphertext, $k$ is a key, and $m$ is the plaintext message. Here, $m$, $k$, and $c$ are sequences of symbols $a$ drawn from alphabet $A^l$ (which has length $l$). Our objective is to train a neural network with parameters $\theta$ to make the approximation $\hat P_{NN}(k, m, \theta) \approx P(k, m)$ such that
\begin{equation}
\label{eqn:objective}
\hat \theta = \arg\min_{\mathbf{\theta}} \mathcal{L}(\hat P_{NN}(k, m, \theta) - P(k, c))
\end{equation}
where $\mathcal{L}$ is L2 loss. We chose this loss function because it penalizes outliers more severely than other loss functions such as L1. Minimizing outliers is important as the model converges to high accuracies (e.g. 95$^+$\%) and errors become infrequent. In this equation, $P(k, c)$ is a one-hot vector and $\hat P_{NN}$ is a real-valued softmax distribution over the same space.
\\

\textbf{Representing ciphers.}
In this work, we chose $A^{27}$ as the uppercase Roman alphabet plus the null symbol, \texttt{'-'}. We encode each symbol $a_i$ as a one-hot vector. The sequences $m$, $k$, and $c$ then become matrices with a time dimension and a one-hot dimension (see Figure \ref{fig:data}). We allow the key length to vary between 1 and 6 but choose a standard length of 6 for $k$ and pad extra indices with the null symbol. If $N$ is the number of time steps to unroll the LSTM during training, then $m$ and $c$ are $N-6 \times 27$ matrices and $k$ is a $6 \times 27$ matrix. To construct training example $(x_i, y_i)$, we concatenate the key, plaintext, and ciphertext matrices according to Equations \ref{eqn:train-example-x} and \ref{eqn:train-example-y}.
\begin{align}
\label{eqn:train-example-x}
x_i = (k, c) \\
\label{eqn:train-example-y}
y_i = (k, m)
\end{align}
Concretely, for the Autokey cipher, we might obtain the following sequences:

\begin{align*}
\textrm{input :  \texttt{KEY---ISSIBIGACPUWGNRBTBBBO}}\\
\textrm{target:  \texttt{KEY---YOUKNOWNOTHINGJONSNOW}}
\label{code:example-text}
\end{align*}

We could also have fed the model the entire key at each time step. However, this would have increased the size of the input layer (and, as a result, total size of $\theta$) introducing an unnecessary computational cost. We found that the LSTM could store the keyphrase in its memory cell without difficulty so we chose the simple concatenation method of Equations \ref{eqn:train-example-x} and \ref{eqn:train-example-y} instead. We found empirical benefit in appending the keyphrase to the target sequence; loss decreased more rapidly early in training.

\section{Our model}

\textbf{Recurrent neural networks (RNNs).} The simplest RNN cell takes as input two hidden state vectors: one from the previous time step and one from the previous layer of the network. Using indices $t=1 \dots T$ for time and $l=1 \dots L$ for depth, we label them $h^l_{t-1}$ and $h^{l - 1}_t$ respectively. Using the notation of Karpathy et al. \cite{Karpathy2016VisualizingNetworks}, the RNN update rule is

\begin{align*}
h^l_t = \tanh W^l \begin{pmatrix}h^{l - 1}_t\\h^l_{t-1}\end{pmatrix}
\end{align*}

where $h \in \mathbb{R}^n$, $W^l$ is a $n \times 2n $ parameter matrix, and the $\tanh$ is applied elementwise.

The Long Short-Term Memory (LSTM) cell is a variation of the RNN cell which is easier to train in practice \cite{Hochreiter1997LongMemory}. In addition to the hidden state vector, LSTMs maintain a memory vector, $c_t^l$. At each time step, the LSTM can choose to read from, write to, or reset the cell using three gating mechanisms. The LSTM update takes the form:

\begin{minipage}{.5\linewidth}
\begin{align*}
&\begin{pmatrix}i\\f\\o\\g\end{pmatrix} =
\begin{pmatrix}\mathrm{sigm}\\\mathrm{sigm}\\\mathrm{sigm}\\\tanh\end{pmatrix}
W^l \begin{pmatrix}h^{l - 1}_t\\h^l_{t-1}\end{pmatrix}
\end{align*}
\end{minipage}%
\begin{minipage}{.5\linewidth}
\begin{align*}
&c^l_t = f \odot c^l_{t-1} + i \odot g\\
&h^l_t = o \odot \tanh(c^l_t)
\end{align*}
\end{minipage}

The sigmoid ($\mathrm{sigm}$) and $\tanh$ functions are applied element-wise, and $W^l$ is a $4n \times 2n$ matrix. The three gate vectors $i,f,o \in \mathbb{R}^n$ control whether the memory is updated, reset to zero, or its local state is revealed in the hidden vector, respectively. The entire cell is differentiable, and the three gating functions reduce the problem of vanishing gradients \cite{Bengio1994LearningDifficult} \cite{Hochreiter1997LongMemory}.
\\

We used a single LSTM cell capped with a fully-connected softmax layer for all experiments (Figure \ref{fig:model}). We also experimented with two and three stacked LSTM layers and additional fully connected layers between the input and LSTM layers, but these architectures learned too slowly or not at all. In our experience, the simplest architecture worked best.

\begin{figure*}
\centering
\begin{subfigure}{\columnwidth}
  \centering
  \includegraphics[width=\columnwidth]{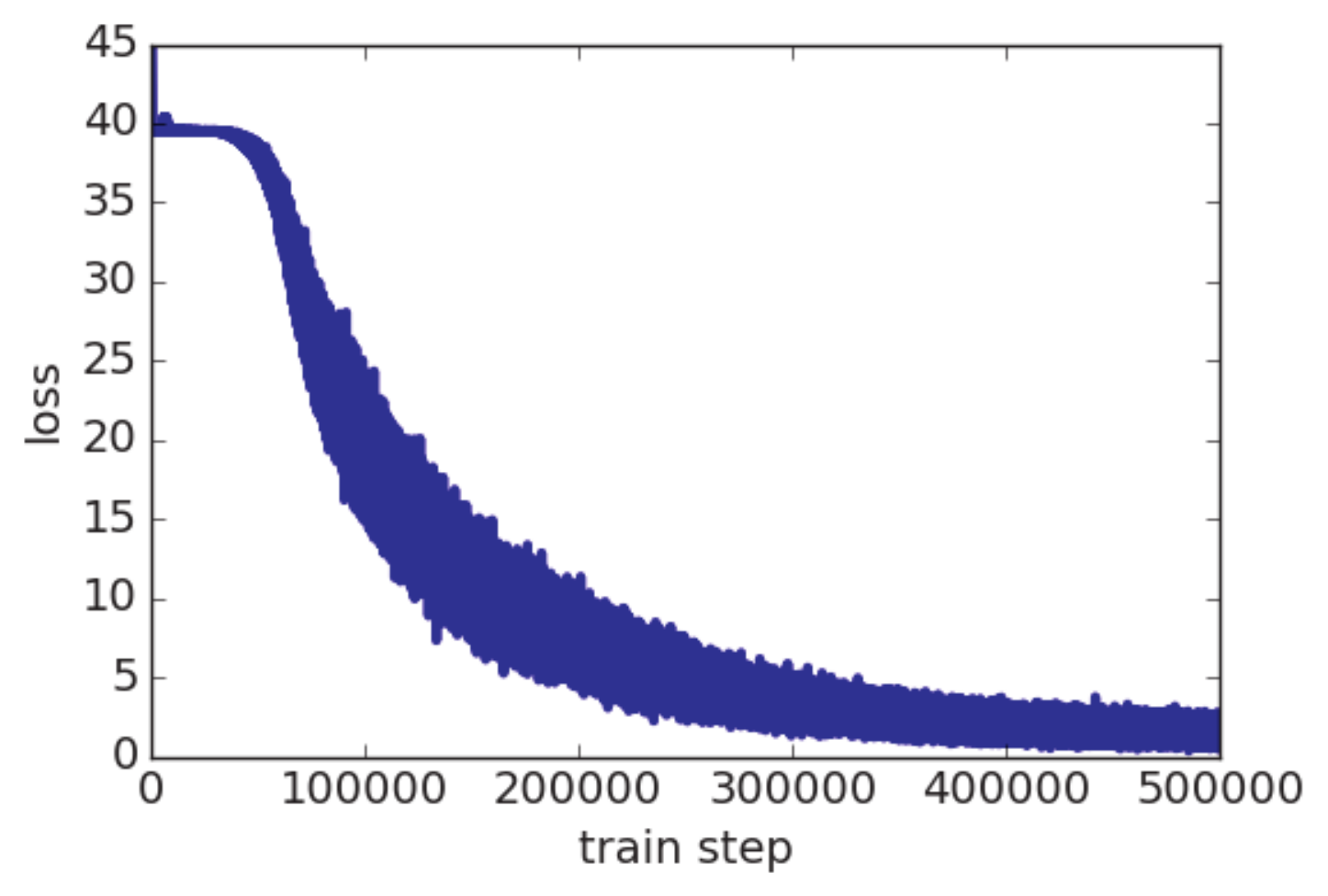}
  \caption{Train loss}
  \label{fig:loss-enigma}
\end{subfigure}
\begin{subfigure}{\columnwidth}
  \centering
  \includegraphics[width=\columnwidth]{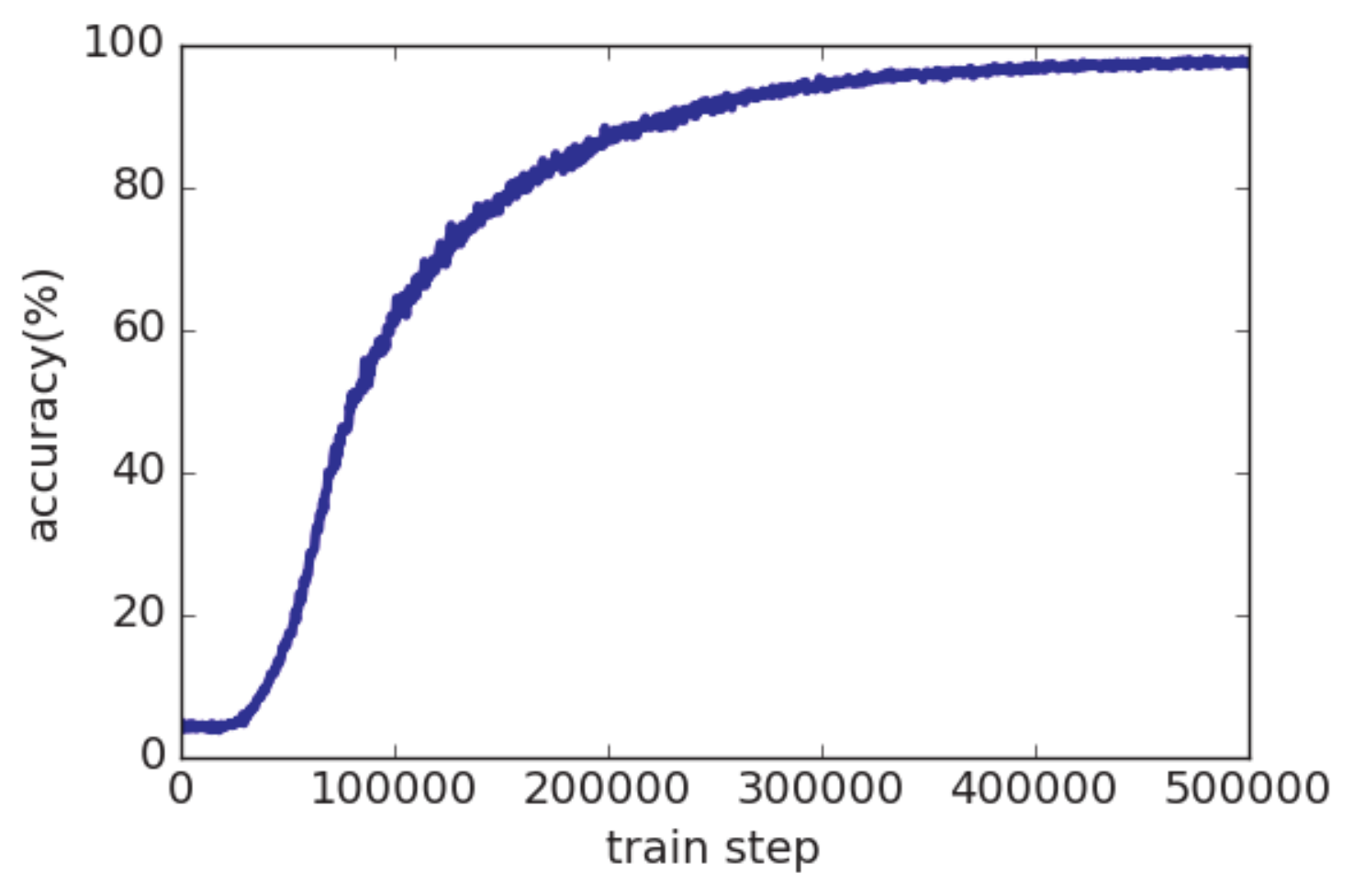}
  \caption{Test accuracy}
  \label{fig:acc-enigma}
\end{subfigure}
\caption{Loss decreases rapidly at first, around 5000 train steps, as the network learns to capture simple statistical distributions. Later, around 100000 train steps, model learns the Enigma cipher itself and accuracy spikes. A significant portion of training, starting around 350000 train steps, is spent gaining the last $5\%$ accuracy.}
\label{fig:train-enigma}
\end{figure*}

\section{Experiments}

We considered three types of polyalphabetic substitution ciphers: the Vigenere, Autokey, and Enigma. Each of these ciphers is composed of various rotations of $A$, called Caesar shifts. A one-unit Caesar shift would perform the rotation $A[i] \mapsto A[(i+1)\mod N]$.

The Vigenere cipher performs Caesar shifts with distances that correspond to the indices of a repeated keyphrase. For a keyphrase $k$ of length $z$, the Vigenere cipher decrypts a plaintext message $m$ according to:

\begin{equation}
P(k^i_t, m^j_t) = A[(j+(i\mod z))\mod l]
\end{equation}

\noindent
where the lower indices of $k$ and $m$ correspond to time and the upper indices correspond to the index of the symbol in alphabet $A^l$.

The Autokey cipher is a variant of this idea. Instead of repeating the key, it concatenates the plaintext message to the end of the key, effectively creating a new and non-repetitive key:

\begin{equation}
i \quad \textrm{as in} \quad
\begin{cases}
    k^i_t, & \text{if } t\leq z\\
     m^i_{t-z}, & \text{otherwise}
\end{cases}
\end{equation}

The Enigma also performs rotations over $A^l$, but with a function $P$ that is far more complex. The version used in World War II contained 3 wheels (each with 26 ring settings), up to 10 plugs, and a reflector, giving it over $1.07 \times 10^{23}$ possible configurations \cite{Miller2017TheEnigma}. We selected a constant rotor configuration of \texttt{A-I-II-III}, ring configuration of \texttt{2, 14, 8} and no plugs. For an explanation of these settings, see \cite{Rejewski1981HowCipher}. We set the rotors randomly according to a 3-character key, giving a subspace of settings that contained $26^{3}$ possible mappings. With these settings, our model required several days of compute time on a Tesla k80 GPU. The rotor and ring configurations could also be allowed to vary by appending them to the keyphrase. We tried this, but convergence was too slow given our limited computational resources.
\\

\textbf{Synthesizing data.} The runtime of $P(k,m)$ was small for all ciphers so we chose to synthesize training data on-the-fly, eliminating the need to synthesize and store large datasets. This also reduced the likelihood of overfitting. We implemented our own Vigenere and Autokey ciphers and used the historically accurate \texttt{crypto-enigma}\footnote{\texttt{http://pypi.python.org/pypi/crypto-enigma}} Python package as our Enigma simulator.

The symbols of the input sequences $k$ and $m$ were drawn randomly (with uniform probability) from the Roman alphabet, $A^{26}$. We chose characters with uniform distribution to make the task more difficult. Had we used grammatical (e.g. English) text, our model could have improved its performance on the task simply by memorizing common n-grams. Differentiating between gains in accuracy due to learning the Enigma versus those due to learning the statistical distributions of English would have been difficult.
\\

\textbf{Optimization.} Following previous work by Xavier et al. \cite{Glorot2010UnderstandingNetworks}, we use the 'Xavier' initialization for all parameters. We use mini-batch stochastic gradient descent with batch size 50 and select parameter-wise learning rates using Adam \cite{Kingma2014Adam:Optimization} set to a base learning rate of $5 \times 10^{-4}$ and $\beta=(0.9, 0.999)$. We trained on ciphertext sequences of length 14 and keyphrase sequences of length 6 for all tasks. As our Enigma configuration accepted only 3-character keys, we padded the last three entries with the null character, \texttt{'-'}.

On the Enigma task, we found that our model's LSTM required a memory size of at least 2048 units. The model converged too slowly or not at all for smaller memory sizes and multi-cell architectures. We performed experiments on a model with 3000 hidden units because it converged approximately twice as quickly as the model with 2048 units. For the Vigenere and and Autokey ciphers, we explored hidden memory sizes of 32, 64, 128, 256, and 512. For each cipher, we halted training after our model achieved $99^+\%$ accuracy; this occurred around $5 \times 10^{5}$ train steps on the Enigma task and $1 \times 10^{5}$ train steps on the others.

The number of possible training examples, $\approx 10^{19}$, far exceeded the total number of examples the model encountered during training, $\approx 10^7$ (each of these examples were generated on-the-fly). We were, however, concerned with another type of overfitting. On the Enigma task, our model contained over thirty million learnable parameters and encountered each three-letter key hundreds of times during training. Hence, it was possible that the model might memorize the mappings from ciphertext to plaintext for all possible ($26^3$) keys. To check for this sort of overfitting, we withheld a single key, \texttt{KEY}, from training and inspected our trained model's performance on a message encrypted with this key.
\\

\begin{figure}
\centering
\includegraphics[width=\columnwidth]{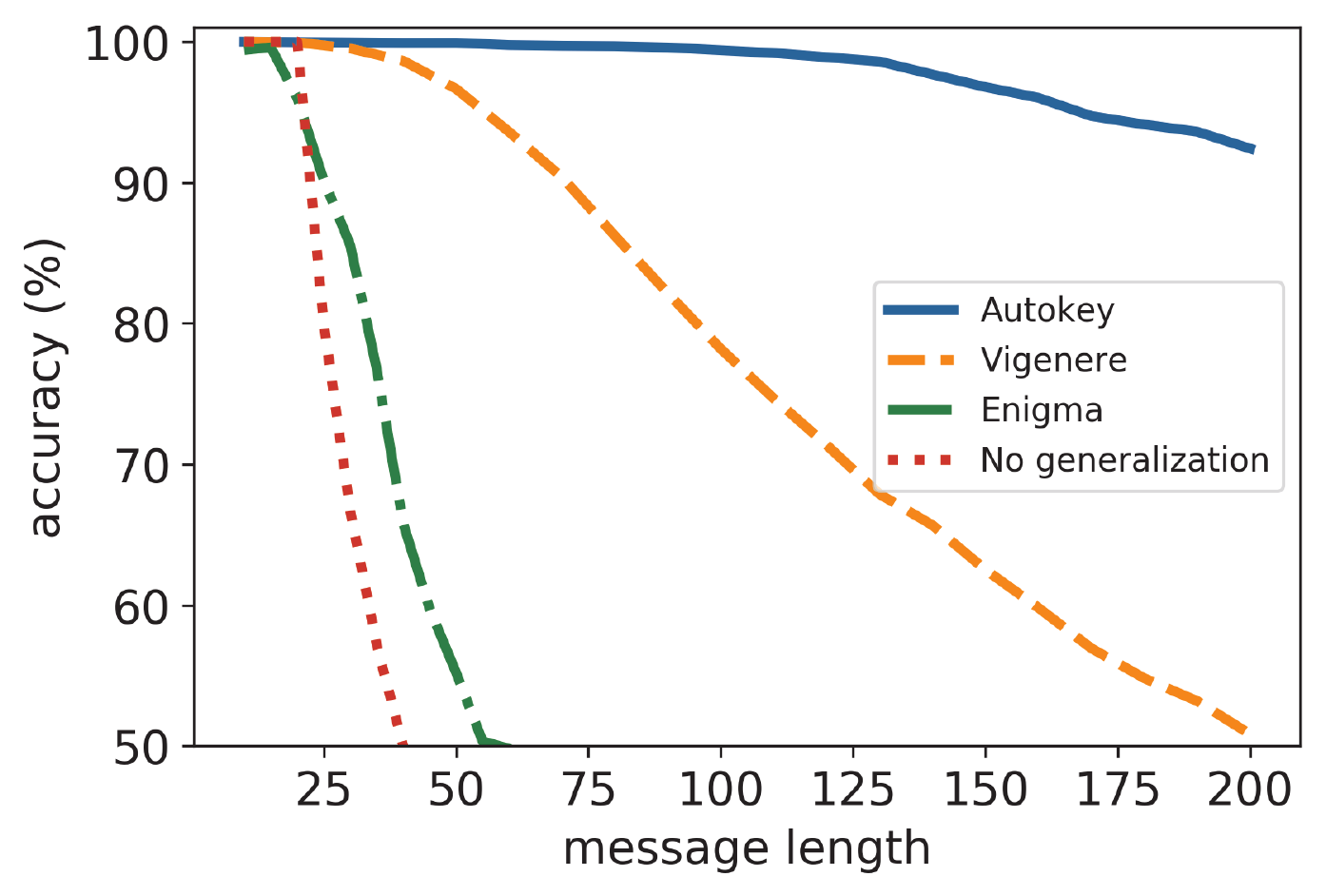}
\caption{The model, trained on messages of 20 characters, generalizes well to messages of over 100 characters for the Vigenere and Autokey ciphers. Generalization occurs on the Enigma, but to a lesser degree, as the task is far more complex.}
\label{fig:generalize-vigenere}
\end{figure}

\textbf{Generalization.} We evaluated out model on two metrics for generalization. First, we tested its ability to decrypt 20 randomly-generated test messages using an unseen keyphrase ('\texttt{KEY}'). It passed this test on all three ciphers. Second, we measured its ability to decrypt messages longer than those of the training set. On this test, our model exhibited some generalization for all three ciphers but performed particularly well on the Vigenere task, decoding sequences an order of magnitude longer than those of the trainign set (see Figure \ref{fig:generalize-vigenere}). We observed that the norm of our model's memory vector increased linearly with the number of time steps, leading us to hypothesize that the magnitudes of some hidden activations increase linearly over time. This phenomenon is probably responsible for reduced decryption accuracy on very long sequences.
\\

\begin{figure}
\centering
\begin{subfigure}{\columnwidth}
  \centering
  \includegraphics[width=\columnwidth]{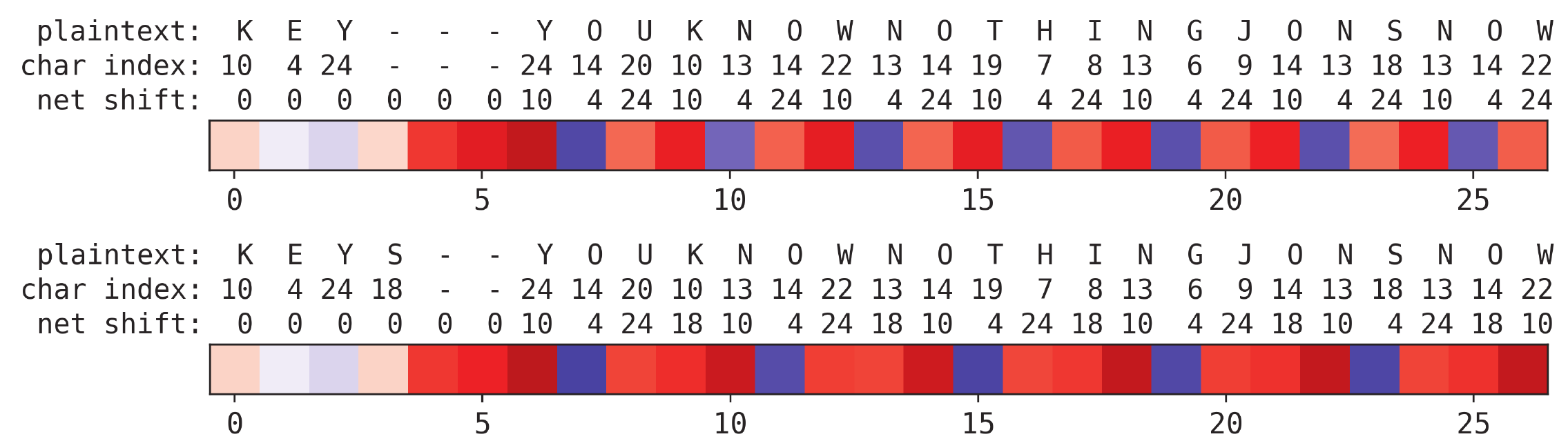}
  \caption{Vigenere (hidden unit 252)}
  \label{fig:neuron-vigenere}
\end{subfigure}\\
\begin{subfigure}{\columnwidth}
  \centering
  \includegraphics[width=\columnwidth]{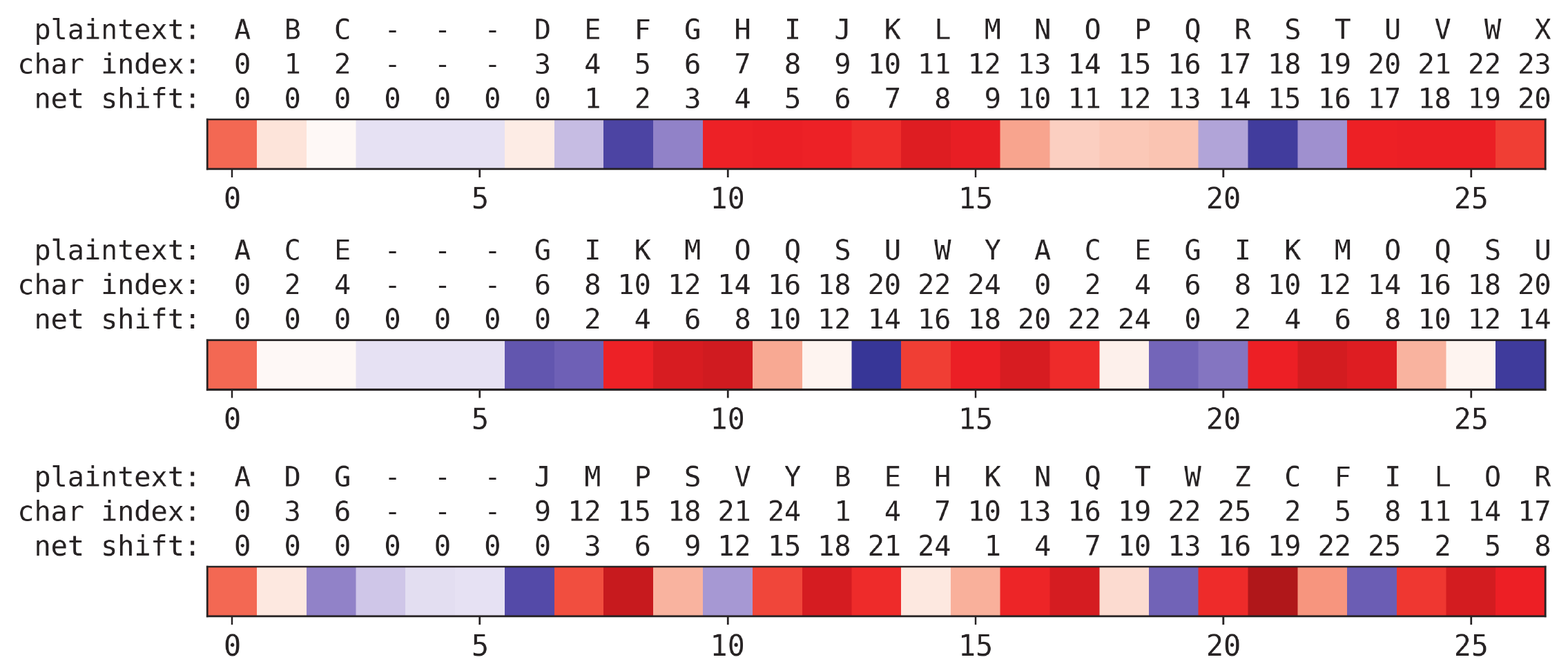}
  \caption{Autokey (hidden unit 30)}
  \label{fig:neuron-autokey}
\end{subfigure}
\begin{subfigure}{\columnwidth}
  \centering
  \includegraphics[width=\columnwidth]{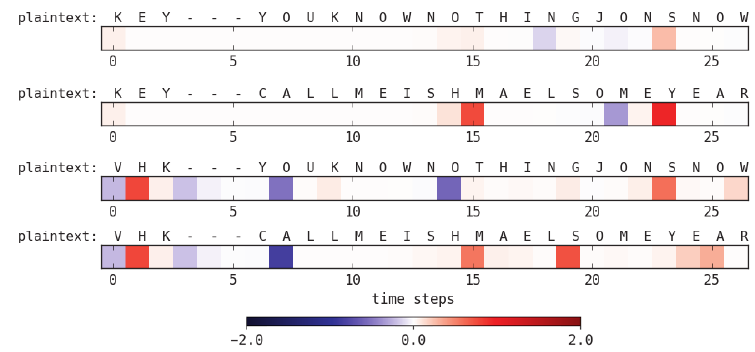}
  \caption{Enigma (hidden unit 1914)}
  \label{fig:neuron-enigma}
\end{subfigure}
\caption{Shown above are examples of plaintext messages decrypted by our model. The red and blue heat maps correspond to the activations of indicated hidden units taken from the LSTM's memory vector $c$. The \texttt{char index} label corresponds to the index of the plaintext character in alphabet $A$. The \texttt{net shift} label corresponds to the number of Caesar shifts between the ciphertext and the plaintext. \\ \\ (a) Hidden unit 252 of the Vigenere model has a negative activation once every $n$ steps where $n$ is the length of the keyphrase (examples for $n=3,4$ are shown). We hypothesize that this is a \textit{timing unit} which allows the model to index into the keyphrase as a function of the encryption step. \\ \\ (b) The 30th hidden unit of the Autokey model has negative activations for specific character indices (e.g. 2 and 14) and positive activations for others (e.g. 6 and 18). We hypothesize that this \textit{shift unit} helps the model compute the magnitude of the Caesar shift between the ciphertext and plaintext. \\ \\ (c) The hidden activations of the Enigma model were generally sparse. Hidden unit 1914 is no exception. For different messages, only its activation magnitude changes. For different keyphrases, its entire activation pattern (signs and magnitudes) change. We hypothesize that it is a \textit{switch unit} which activates only when the Enigma enters a particular rotor configuration.}
\label{fig:neurons}
\end{figure}

%
\begin{figure*}
\centering
\begin{subfigure}{\columnwidth}
  \centering
  \includegraphics[width=\columnwidth]{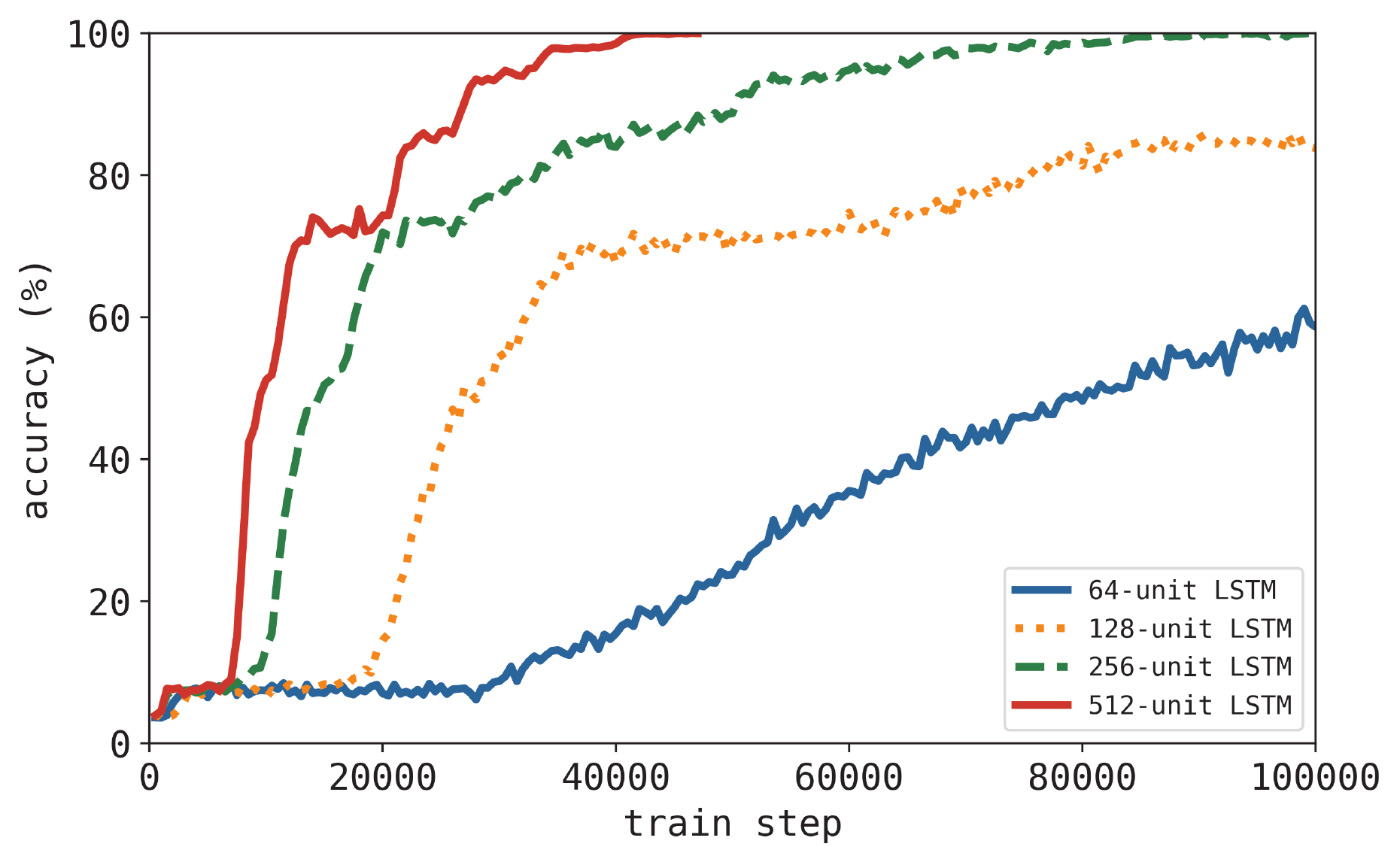}
  \caption{Vigenere task}
  \label{fig:mem-vig}
\end{subfigure}
\begin{subfigure}{\columnwidth}
  \centering
  \includegraphics[width=\columnwidth]{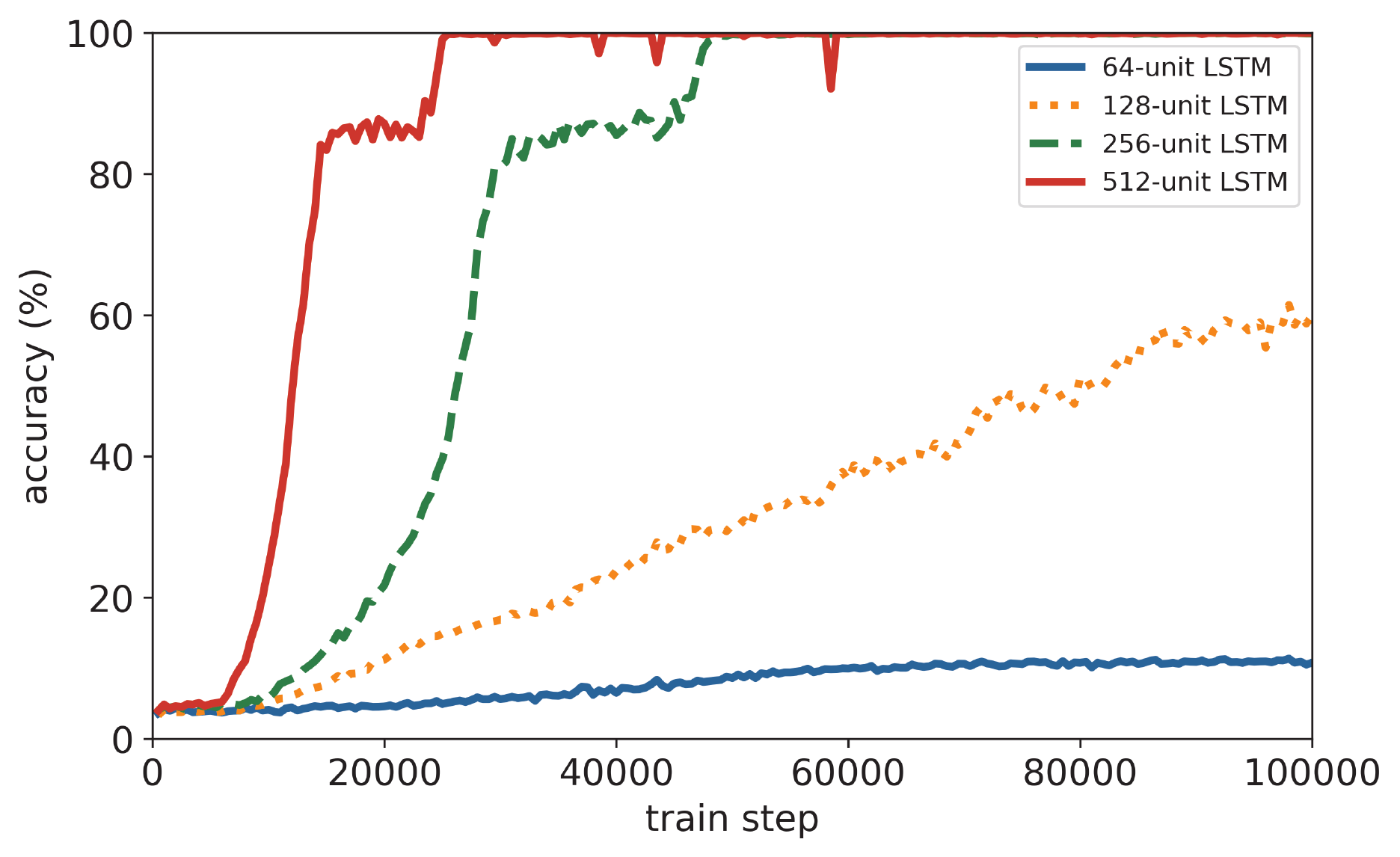}
  \caption{Autokey task}
  \label{fig:mem-auto}
\end{subfigure}
\caption{Shown above are test accuracies of our model on the Vigenere and Autokey cipher tasks. Notice that for small RNN memory sizes (64 and 128 hidden units), the model achieves better performance on the Vigenere task. Meanwhile, for large memory sizes (256 and 512 hidden units), the model converges to 99$^+\%$ accuracy more rapidly on the Autokey task. Evidently, the model's test accuracy is more sensitive to memory size on the Autokey task than on the Vigenere task.}
\label{fig:memory}
\end{figure*}

\textbf{Memory usage.} Based on our model's ability to generalize over unseen keyphrases and message lengths, we hypothesized that it learns an efficient internal representation of each cipher. To explore this idea, we first examined how activations of various units in the LSTM's memory vector changed over time. We found, as shown in Figure \ref{fig:neurons}, that 1) these activations mirrored qualitative properties of the ciphers and 2) they varied considerably between the three ciphers.

We were also interested in \textit{how much} memory our model required to learn each cipher. Early in this work, we observed that the model required a very large memory vector (at least 2048 hidden units) to master the Enigma task. In subsequent experiments (Figure \ref{fig:memory}), we found that the size of the LSTM's memory vector was more important when training on the Autokey task than the Vigenere task. We hypothesize that this is because the model must continually update its internal representation of the keyphrase during the Autokey task, whereas on the Vigenere task it needs only store a static representation of the keyphrase. The Enigma, of course, requires dramatically more memory because it must store the configurations of three 26-character wheels, each of which may rotate at a given time step.

Based on these observations, we claim that the amount of memory our model requires to learn a given cipher can serve as an informal measure of how much each encryption step depends on previous encryption steps. When characterizing a black box cipher, this information may be of interest.
\\

\textbf{Reconstructing keyphrases.} Having verified that our model learned internal representations of the three ciphers, we decided to take this property one step further. We trained our model to predict keyphrases as a function of plaintext and ciphertext inputs. We used the same model architecture as described earlier, but the input at each timestep became two concatenated one-hot vectors (one corresponding to the plaintext and one to the ciphertext). Reconstructing the keyphrase for a Vigenere cipher with known keylength is trivial: the task reduces to measuring the shifts between the plaintext and ciphertext characters. We made this task more difficult by training the model on target keyphrases of unknown length (1-6 characters long). In most real-world cryptanalysis applications, the length of the keyphrase is unknown which also motivated our choice. Our model obtained $99^+\%$ accuracy on the task.

Reconstructing the keyphrase of the Autokey was a more difficult task, as the keyphrase is used only once. This happens during the first 1-6 steps of encryption. On this task, accuracy exceeded 95\%. In future work, we hope to reconstruct the keyphrase of the Engima from plaintext and ciphertext inputs.
\\

\textbf{Limitations.} Our model is data inefficient; it requires at least a million training examples to learn a cipher. If we were interested in characterizing an unknown cipher in the real world, we would not generally have access to unlimited examples of plaintext and ciphertext pairs.

Most modern encryption techniques rely on public-key algorithms such as RSA. Learning these functions requires multiplying and taking the modulus of large numbers. These algorithmic capabilities are well beyond the scope of our model. Better machine learning models may be able to learn simplified versions of RSA, but even these would probably be data- and computation-inefficient.

\section{Conclusions}

This work proposes a fully-differentiable RNN model for learning the decoding functions of polyalphabetic ciphers. We show that the model can achieve high accuracy on several ciphers, including the the challenging 3-wheel Enigma cipher. Furthermore, we show that it learns a general algorithmic representation of these ciphers and can perform well on 1) unseen keyphrases and 2) messages of variable length.

%
Our work represents the first general method for reconstructing the functions of polyalphabetic ciphers. The process is fully automated and an inspection of trained models offers a wealth of information about the unknown cipher. Our findings are generally applicable to analyzing any black box sequence-to-sequence translation task where the translation function is a deterministic algorithm. Finally, decoding Enigma messages is a complicated algorithmic task, and thus we suspect learning this task with an RNN will be of general interest to the machine learning community.

\section{Acknowledgements}
We are grateful to Dartmouth College for providing access to its cluster of Tesla K80 GPUs. We thank Jason Yosinski and Alan Fern for insightful feedback on preliminary drafts.

\bibliography{crypto-rnn}
\bibliographystyle{aaai}

\end{document}